%% file: main.tex
\documentclass[pmlr]{jmlr}

\RequirePackage{graphicx}
\usepackage{booktabs}
\usepackage{longtable}
\makeatletter
\def\set@curr@file#1{\def\@curr@file{#1}}
\makeatother
\usepackage{siunitx}
\usepackage{enumitem}
\usepackage{tikz}
\usepackage{float}
\DeclareUnicodeCharacter{2265}{\ensuremath{\geq}}
\theorembodyfont{\upshape}
\theoremheaderfont{\scshape}
\theorempostheader{:}
\theoremsep{\newline}

\jmlrvolume{[VOLUME \# TBD]}
\jmlryear{2025}
\jmlrworkshop{Machine Learning for Healthcare}
\newcommand{\cut}[1]{}

\title[Enhancing Adaptive Behavioral Intervention Policies]
{Enhancing Adaptive Behavioral Interventions with LLM Inference from Participant-Described States}

\author{\Name{Karine Karine}
       \Email{karine@cs.umass.edu}\\ 
       \addr University of Massachusetts\\
       Amherst, MA, USA 
       \AND
       \Name{Benjamin M. Marlin}
       \Email{marlin@cs.umass.edu}\\ 
       \addr University of Massachusetts\\
       Amherst, MA, USA} 

\begin{document}
\maketitle
\begin{abstract}

\input{LLM_TS0_abstract}
\end{abstract}
\input{LLM_TS1_introduction}
\input{LLM_TS2_background}

\input{LLM_TS3_methods}

\input{LLM_TS4_experiments_setup}

\input{LLM_TS5_conclusion}

\section*{Acknowledgments}
This work is supported by National Institutes of Health National Cancer Institute, Office of Behavior and Social Sciences, and National Institute of Biomedical Imaging and Bioengineering through grants U01CA229445, 1P41EB028242 and P30AG073107.

\bibliography{LLM_TS6_source}
\appendix
\input{LLM_TS7_suppA}
\end{document}

%% file: LLM_TS0_abstract.tex
The use of reinforcement learning (RL) methods to support health behavior change via personalized and just-in-time adaptive interventions is of significant interest to health and behavioral science researchers focused on problems such as smoking cessation support and physical activity promotion. However, RL methods are often applied to these domains using a small collection of context variables to mitigate the significant data scarcity issues that arise from practical limitations on the design of adaptive intervention trials.
In this paper, we explore an approach to significantly expanding the state space of an adaptive intervention without impacting data efficiency. The proposed approach enables intervention participants to provide natural language descriptions of aspects of their current state. It then leverages inference with pre-trained large language models (LLMs) to better align the policy of a base RL method with these state descriptions. To evaluate our method, we develop a novel physical activity intervention simulation environment that generates text-based state descriptions conditioned on latent state variables using an auxiliary LLM.
We show that this approach has the potential to significantly improve the performance of online policy learning methods.

%% file: LLM_TS1_introduction.tex
\section{Introduction}
\label{sec: LLM4TS Introduction}

The use of reinforcement learning (RL) methods \citep{sutton1998reinforcement} to support health behavior change via personalized and just-in-time adaptive interventions is of significant interest to health and behavioral science researchers focused on problems such as smoking cessation support and physical activity promotion.  \citep{coronato2020reinforcement, liao2020personalized, gonul2021reinforcement, yu2021reinforcement}. However, in the adaptive behavioral intervention domain, RL methods are often applied using a small collections of affective, behavioral, physiological and/or environmental context variables to mitigate the significant data scarcity issues that arise from practical limitations on adaptive intervention trials. This includes limited numbers of intervention opportunities per day, limited numbers of study participants, and limited overall study durations.

An important consequence of the use of small state spaces to mitigate data scarcity issues is that the resulting RL policies have an extremely narrow view of the overall state of an intervention participant's health and well-being. This can result in an adaptive intervention system recommending intervention options that range from sub-optimal to inappropriate with respect to the participant's overall state. For example, a physical activity adaptive intervention that conditions only on location, weather, temperature and recent activity level would have no ability to account for the fact that the participant has the flu or has sprained their ankle and cannot walk. Continuing to issue intervention content that disregards the overall health state of a participant may needlessly contribute to increasing habituation \citep{dimitrijevic1972habituation,liao-ubi2018} as well as risk of disengagement from the intervention \citep{park2023understanding}. 

\begin{figure}[t!]
  \begin{minipage}[c]{0.55\textwidth}
    \caption{LLM4TS is a hybrid method that combines LLM inference with a base RL method to improve action selection. The RL agent proposes a candidate action $\tilde{a}_t$. A pre-trained LLM is then used to infer whether the action is aligned with a participant-provided state description. The LLM inference step uses a prompt with multiple components including questions that guide chain of thought-like reasoning.}
    \label{fig: Overview}
  \end{minipage}
  \hfill
  \begin{minipage}[c]{0.40\textwidth}
    \includegraphics[width=\columnwidth]{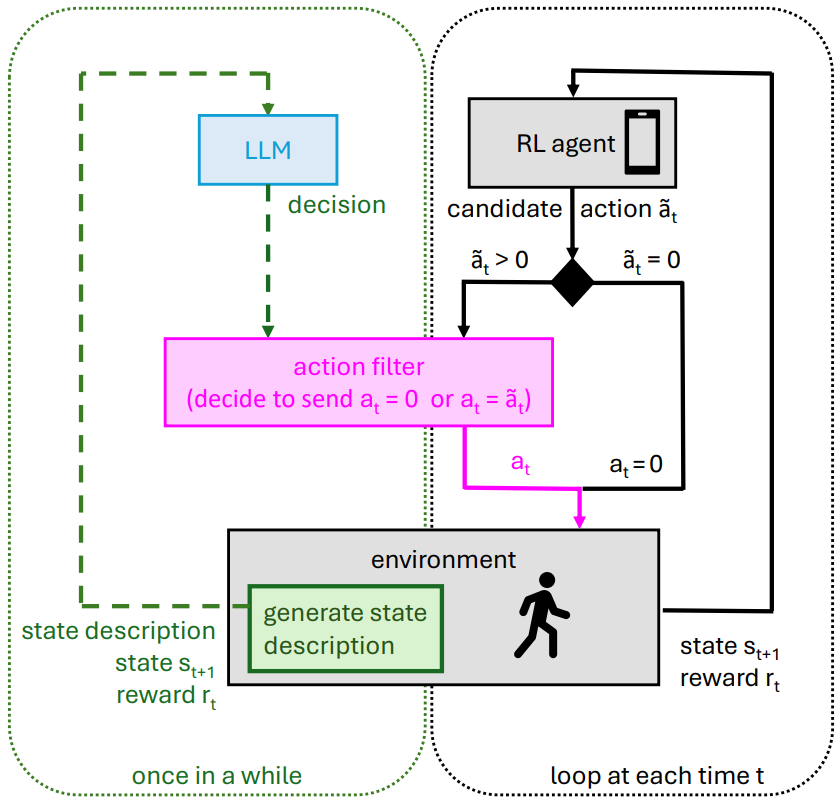}
  \end{minipage}
\end{figure}

\cut{
\begin{figure}[!t]
\centering
\includegraphics[width=0.4\columnwidth]{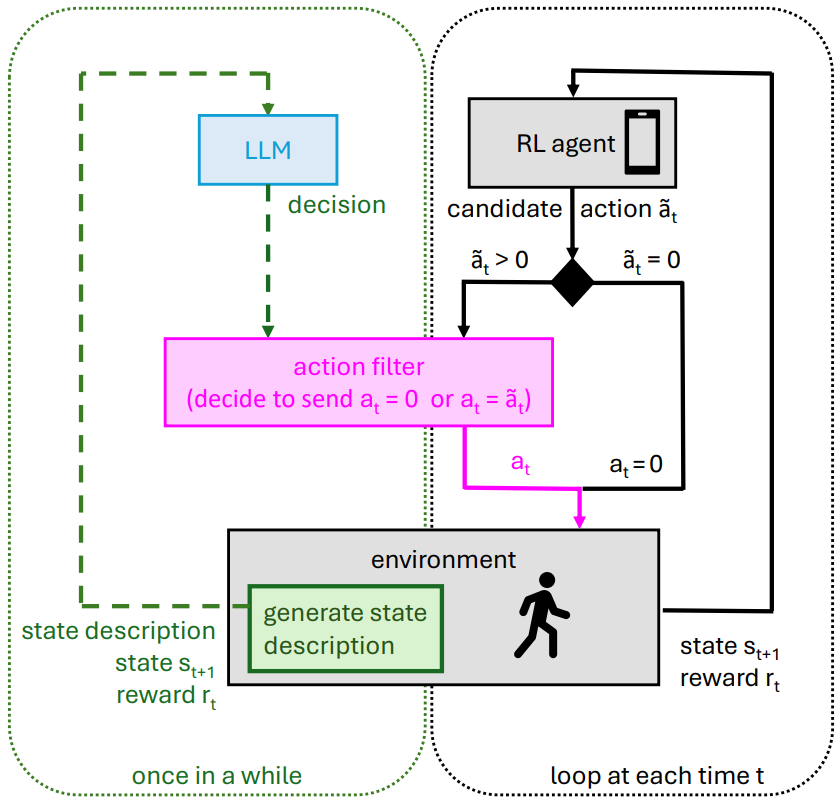}
\caption{Overview: LLM4TS is a hybrid method that combines LLM inference with a base RL method to improve action selection. The RL agent proposes a candidate action $\tilde{a}_t$. Inference using a pre-trained LLM is then used to determine whether the selected action is well aligned with a participant-provided state description. The LLM inference step uses a prompt with multiple components including a description of the behavioral dynamics, the state description, and questions that guide chain of thought-like reasoning. 
\vspace{-2em}}
\label{fig: Overview}
\end{figure}
}

In this paper, we propose to leverage the natural language understanding and reasoning capabilities of pre-trained large language models (LLMs) \citep{vaswani2017attention, achiam2023gpt, grattafiori2024llama} to help mitigate the problem of restricted state spaces in RL-based adaptive interventions without impacting data efficiency or compromising the ability of behavioral science researchers to control intervention content. The approach that we explore is based on (1) enabling intervention participants to describe any aspect of their state in free text, (2) using a data-efficient RL algorithm with limited state to propose a candidate action, and (3) applying an LLM with a specifically engineered prompt to perform inference with the goal of deciding whether the proposed action is aligned with the participant's most recently declared state. We use Thompson sampling as a data-efficient base RL algorithm in this work \citep{russo2018,  chu2011contextual, Thompson1933}. We refer to the resulting method as LLM4TS following the taxonomy of \cite{pternea2024rl}. We provide an overview of our method in Figure \ref{fig: Overview}.

While the continuous collection of state descriptions from participants would be burdensome, collecting this data continuously is not required under the proposed approach. We conceptualize the process of participants describing relevant aspects of their state as providing a response to a daily query such as asking them to describe how they are feeling each morning or to indicate if they are experiencing any barriers to participating in the intervention. This interaction could be supplemented with the ability for a participant to provide additional descriptions via an on-demand interaction in response to events or other changes in health state. Further, while we use text directly in this work, current speech-to-text capabilities would easily allow participants to supply this information using voice, further reducing burden \citep{radford2023robust,kuhn2024measuring}.  

To evaluate our approach, we build on a recently introduced simulation environment that is based on an adaptive messaging intervention for physical activity promotion. This simulation includes models of key aspects of behavioral intervention dynamics including intervention habituation and disengagement risk \citep{karine2024}. We add to this system a simulation of participants responding to the morning query about their general health state. We generate the responses using an LLM prompt that conditions on latent dimensions of the true underlying health state of the simulated participant. In this work, we focus on a latent binary state indicating whether the participant is able to engage in physical activity (specifically, walking) or not.

We present extensive results showing that LLMs can be used to reliably generate varied state descriptions when conditioned on underlying state variables, that LLMs can accurately infer when to filter actions based on state descriptions, and that the proposed LLM4TS method results in improved performance relative to the standard TS method when periods where a participant cannot engage in walking are encountered. We further explore the impact of different components of the LLM inference prompt as well as run-time metrics for the LLM inference process. 

\vspace{1em}
    
\noindent \textbf{The primary contributions of this work are}:

\begin{enumerate}[leftmargin=*]

    \item \textbf{LLM4TS.} We introduce an ``LLM as judge'' approach  to enhancing personalized adaptive health interventions. LLM4TS leverages the natural language understanding and reasoning capabilities of LLMs to help mitigate the limited state representation of a Thompson Sampler while maintaining data efficiency. 
    
    \item \textbf{StepCountJITAI+LLM.} We develop a novel simulation environment to evaluate the proposed method. 
    This simulation environment extends an existing base simulator to add support for simulating participant generated text using LLMs. The simulator generates text-based descriptions of participant state that reflect latent state dimensions. This simulation environment has significant potential to enable the development of new RL algorithms tailored to the adaptive intervention domain.
\end{enumerate}


\subsection*{Generalizable Insights about Machine Learning in the Context of Healthcare}

In this work, we develop and evaluate an approach to augmenting a base RL method with LLM-based reasoning capabilities to help address the significant challenge of data scarcity that arises when applying RL methods to optimize adaptive intervention policies in the context of practical research study designs. We focus specifically on leveraging the common sense reasoning capabilities of pre-trained LLMs to better align action selection with participant-generated state descriptions. While the method is evaluated in the context of a physical activity adaptive intervention simulation in this paper, we expect that the advantages demonstrated over the base RL approach will generalize to real physical activity studies, as well as other adaptive intervention problem domains. In summary, we believe this is a promising approach for significantly augmenting the intelligence of adaptive health interventions while respecting practical constraints on study designs and retaining the ability of intervention designers to completely control intervention content. 
  

%% file: LLM_TS2_background.tex
\section{Background and Related Work}
\label{sec: LLM RL Background}

In this section, we describe background on adaptive interventions, intervention policy learning, simulation environments for adaptive intervention research, and other related work. 

\vspace{.5em}

\noindent\textbf{Adaptive Sequential Interventions}. In an adaptive sequential intervention, the overall intervention package consists of multiple intervention \textit{options}. Different intervention options are selected and provided to a patient or study participant at each of a collection of time points referred to as \textit{decision points}. The selection of intervention options at each decision point follows an intervention policy (or decision rule) that takes as input selected aspects of the overall state of the individual \citep{collins2007multiphase}. 

Adaptive sequential intervention designs have been widely studied in the behavioral science and mobile health research communities, resulting in multiple frameworks including the multiphase optimization strategy and sequential multiple assignment randomized trial \citep{collins2007multiphase}, as well as the Just-In-Time Adaptive Intervention (JITAI) \citep{nahum2018just}. In this work, we focus on the JITAI setting, where the goal is often described as providing the right type and amount of support at the right time by adapting to an individual’s changing internal and contextual state \citep{nahum2018just}. 

JITAIs have been developed and studied in areas including physical activity promotion \citep{hardeman2019systematic}, weight loss \citep{forman2019randomized}, diet adherence \citep{goldstein2021optimizing} and tobacco and other substance use \citep{yang2023just,perski2022technology}.  Such interventions target health behaviors that are known to be driving risk factors for multiple chronic illnesses that account for 86\% of all US healthcare spending  \citep{holman2020relation}.

\vspace{0.5em}
\label{Section Background: Thompson Sampling}
\noindent\textbf{Intervention Policy Learning.} Given a set of intervention options, the key problem in adaptive intervention design is mapping the context (or state) of an individual into the selection of an optimal intervention option at each decision point. Since an adaptive intervention is a sequential decision making problem, the natural optimization approaches are control theory methods \citep{golnaraghi2010automatic} and reinforcement learning (RL) \citep{sutton1998reinforcement}. Both approaches have been used to optimize JITAIs in prior work \citep{liao2020personalized,gonul2021reinforcement,Mistiri-jcs2025}.
In this work, we focus on learning intervention policies using RL methods.

The correspondence between adaptive intervention terminology and standard RL terminology is straightforward: the intervention options are the actions, the participant state is the environment state, and measurements of the proximal or distal outcomes of interest forms the basis for the reward \citep{sutton1998reinforcement}. However, the adaptive intervention research study domain is a challenging setting for the application of RL methods due to practical limits on the number of decision points per day (often less than 10), the number of study participants (often 10's to low 100's), and the per-participant study duration (typically several weeks to one year). These limits result in significant data scarcity and require the application of highly data efficient RL methods. 

One approach to achieve data efficiency is the application of low-variance, high-bias RL approaches such as Thompson Sampling (TS). While full RL methods attempt to estimate the future impact of present actions in each state, TS methods essentially select actions to optimize expected immediate rewards in each state. This is often referred to as the \textit{contextual bandit} setting \citep{russo2018,  chu2011contextual, Thompson1933}. While the contextual bandit assumptions are not satisfied in the adaptive intervention optimization setting, contextual bandit methods such as TS tend to outperform full RL methods under significant data scarcity.

The standard linear Gaussian Thompson Sampling algorithm uses a reward model of the form $\mathcal{N}(r;\theta_a^\top \mathit{v}_t,\sigma_{Ya}^2)$, where $\mathit{v}_t$ is the state vector at time $t$, $\theta_a$ is a vector of weights, and $\sigma_{Ya}^2$ is the reward variance for action $a$. Thus, $\theta_a^\top \mathit{v}_t$ represents the mean reward for action $a$. The reward model weights $\theta_a$ are treated as random variables with distribution $\mathcal{N}(\theta_a;\mu_{ta},\Sigma_{ta})$. Actions are selected at each time $t$ by sampling $\hat{\theta}_a$ from $\mathcal{N}(\theta_a;\mu_{ta},\Sigma_{ta})$ for each action $a$ and choosing the action $a$ with the largest value $\hat{\theta}_a^\top \mathit{v}_t$. The prior distribution for $\theta_a$ is of the form $\mathcal{N}(\theta_a;\mu_{0a},\Sigma_{0a})$. The distribution over $\theta_a$ for the selected action is updated at time $t$ based on the observed reward $r_t$ and $\mathit{v}_t$ using Bayesian inference. We provide the update equations for the mean and covariance matrix below. 
\vspace{-.5em}
\begin{align}
    \Sigma_{(t+1)a} &= \sigma_{Ya}^2  ~ \big( \mathit{v}_t^\top   \mathit{v}_t + \sigma_{Ya}^2 ~ \Sigma^{-1}_{ta}  \big)^{-1}\label{equations:TS posterior1}\\
    \mu_{(t+1)a}   &= \Sigma_{(t+1)a}  ~ \big((\sigma_{Ya}^2)^{-1} ~ r_t ~ \mathit{v}_t +  \Sigma^{-1}_{ta} ~ \mu_{ta}  \big)
    \label{equations:TS posterior2}
\end{align}

\vspace{.5em}

\label{sec: background StepCountJITAI}
\noindent\textbf{Adaptive Intervention Simulation Environments.} A further challenge with the application of RL methods in the adaptive intervention setting is the extremely high cost of evaluating methods in the context of real human subjects studies. In other application areas of RL where fielding experimental methods or policies can have high cost, such as robotics, RL methods are typically developed using simulation environments \citep{kim-robots}. However, in the adaptive intervention domain, there is very limited prior work on simulation environments. In this work, we extend the physical activity adaptive intervention simulator introduced by \citet{karine2024} called StepCountJITAI. 

StepCountJITAI was specifically designed to support the development of new RL algorithms for the adaptive behavioral intervention domain. It simulates a messaging-based adaptive physical activity intervention. In this simulation environment, the state includes a binary context variable $c_t\in\{0,1\}$ that can be used to model a time varying binary state such as  `stressed/not stressed' or `at home/not at home,' etc. The simulation also models the dynamics of two key behavioral state variables: habituation level $h_t$ and disengagement risk level $d_t$. The variable $a_t$ denotes the action at time $t$, which corresponds to the choice of intervention option. 
The possible actions $a_t$ are: do not send a message ($a_t=0$), send an untailored message ($a_t=1$), send a message tailored to context $0$ ($a_t=2$) and send a message tailored to context $1$ ($a_t=3$). The goal in this domain is to maximize the participant's total walking step count over the duration of the intervention. Thus, the step count at time $t$ serves as the reward for the action taken at time $t$: $r_t=z_{t}$. Further details of the StepCountJITAI simulator are described in Appendix \ref{Section Background: behavioral dynamics}. In this work, we extend the base StepCountJITAI simulator with additional state variables as well as the ability to produce natural language output describing aspects of the full state.


\label{Appendix: Related work}
\vspace{0.5em}
\noindent\textbf{LLMs Combined with RL.} There has been much recent work on approaches that combine Large Language Models (LLMs) with RL to address different problems \citep{pternea2024rl}. The use of reinforcement learning from human feedback to fine-tune LLMs is likely the most well-known such approach \citep{NEURIPS2022_b1efde53}, but our focus is on the opposite problem of using LLMs to enhance RL methods. \cite{pternea2024rl} present a helpful taxonomy that refers to these two categories of approaches as RL4LLM and LLM4RL. The closest LLM4RL work to our approach leverages LLMs to enhance base RL policies, including work that uses an LLM as a policy prior \citep{pternea2024rl,hu2023language}. Our approach falls into this same category, but instead leverages the LLM to judge proposals output by a base RL policy, an example of the LLM-as-judge framework that has previously been applied as part of the automated evaluation of models such as chat assistants \citep{Zheng2023}. To the best of our knowledge, this paper is the first to study the application of an LLM-as-judge approach in combination with a bandit-based RL method to combat significant RL data scarcity issues. Further, we evaluate LLM prompting frameworks that leverage aspects of intermediate reasoning and provision of domain-specific knowledge to improve LLM inference \citep{Wei2022, Lewis2020}.  


%% file: LLM_TS3_methods.tex
\section{Methods}
\label{sec: LLM RL Methods}

In this section, we describe our proposed method as well as our novel simulation environment. Figure \ref{fig: Overview} provides an overview of the proposed method.

\subsection{Proposed Method: LLM4TS}
We propose a hybrid framework where a base RL agent outputs a candidate action at each time step based on an observed state vector. Then, based on an LLM prompt that includes a participant-provided state description, the LLM decides whether to allow or not allow the candidate RL action. Importantly, we assume that the participant has a true, overall state denoted by $s_t$, but that the base RL method only has access to an observed state vector $\tilde{s}_t$ containing values for a small fixed subset of dimensions of $s_t$. Further, we assume that the participant can separately provide a free text description $f_t$ reflecting aspects of their true state $s_t$. This framework enables an end-to-end adaptive intervention where the common sense reasoning ability of a pre-trained LLM can be applied to interpret the participant-provided state description $f_t$ to gain information about $s_t$ that is not accessible via $\tilde{s}_t$. We describe the steps in the framework in detail below. Without loss of generality, we assume that action $0$ is a null action where no intervention content is provided to the participant. 

\vspace{.5em}

\begin{figure}
\includegraphics[width=6in]{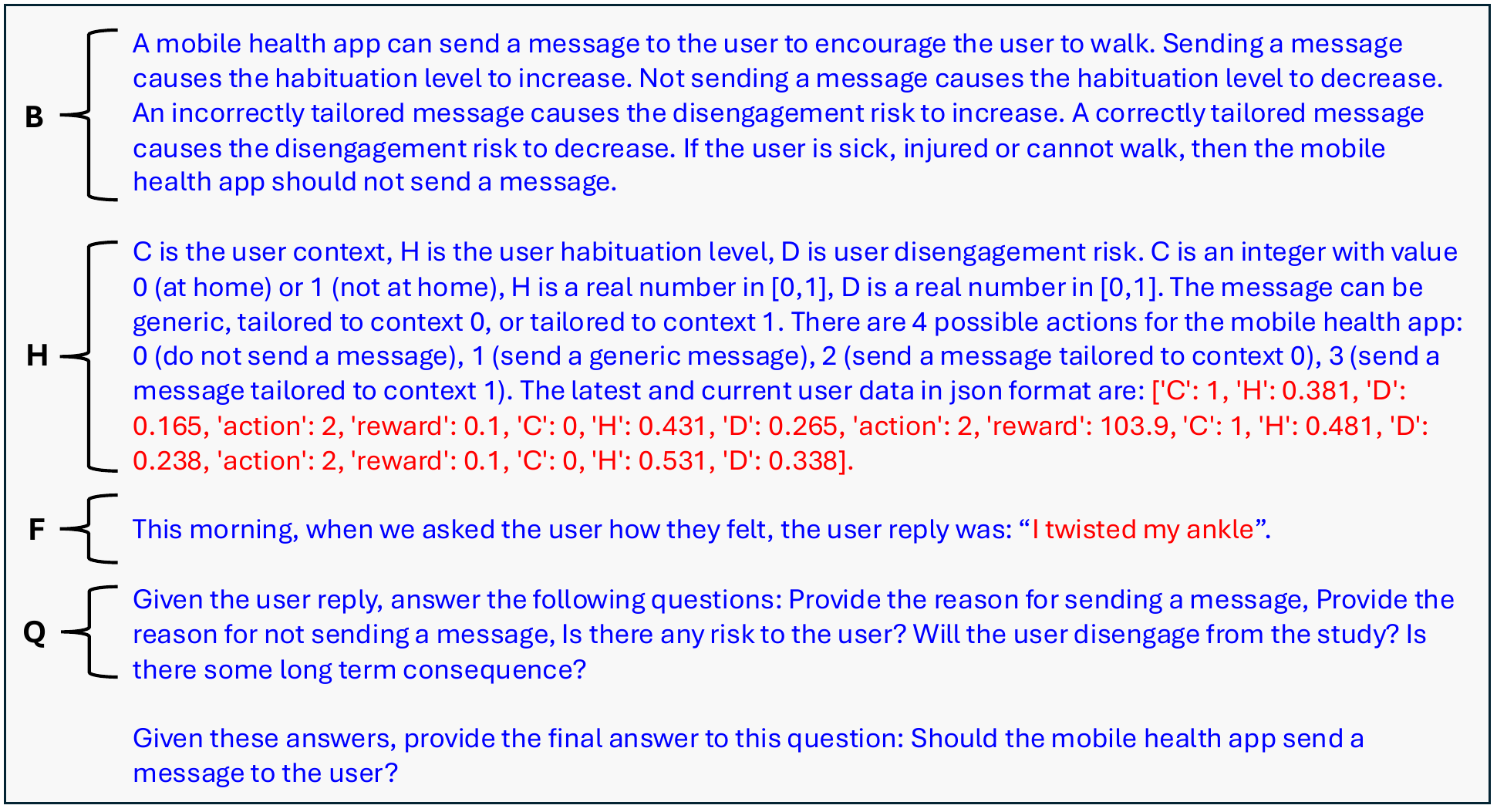}
    \caption{\vspace{-2em} Example LLM prompt template with B, F, Q, H components annotated. Static prompt content is shown in blue. Dynamic prompt content is shown in red. \vspace{-1.5em}}
    \label{fig:prompt1}
\end{figure}

\begin{enumerate}[nosep, leftmargin=*]

    \item \textbf{Candidate Action Generation:} At each time step $t$, the base RL agent proposes a candidate action $\tilde{a}_t$ based on its current parameters $\theta_t$ and the partial state observation $\tilde{s}_t$. If the candidate action is $\tilde{a}_t=0$, the final action is  $a_t=0$. If the candidate action is $\tilde{a}_t\neq 0$, we then apply LLM inference. 

    \item \textbf{LLM Inference:} Given the current participant provided state description $f_t$ and a prompt template, we construct a specific LLM prompt and apply an LLM to perform inference. We then extract the LLM's judgment from the LLM response.

    \item \textbf{Action Filtering:} If the LLM decision is to not allow the base RL method's action, set $a_t = 0$. Otherwise, set $a_t = \tilde{a}_t$.

    \item \textbf{Policy Execution and Update:} Take the action $a_t$. Observe the new partial state $\tilde{s}_{t+1}$ and reward $r_t$. Update the RL agent's parameters based on the tuple $(\tilde{s}_t, a_t, r_t)$, obtaining $\theta_{t+1}$. Query the participant for an updated state description $f_{t+1}$. 
\end{enumerate}

\vspace{.5em}

\label{Section: Construct LLM prompt}
Applying this framework requires specifying intervention options, specifying the observed state space and reward, selecting a pre-trained LLM model, designing the prompt template, and selecting a base RL method. We will describe the intervention options, state space and rewards that we use in our experiments in the next section. We evaluate several LLMs ranging from 3 to 70  billion parameters. We construct the LLM prompt template for our experiments by combining several components: (B) a  description of the adaptive intervention domain and hypothesized behavioral dynamics, (F)  the free text participant provided state description, (Q) intermediate reasoning questions to guide LLM inference, and (H) a short trajectory history consisting of the four most recent (state, action, reward) tuples. Every prompt ends with a final question asking the LLM to make a decision to send the candidate message. We provide an example LLM prompt with all components annotated in Figure \ref{fig:prompt1}. We primarily experiment with the BFQH components when trajectory history information is available and the BFQ components otherwise.  As noted previously, we use Thompson Sampling as the base RL method. 

\cut{
\begin{table}[t!]
    \begin{center}
    \caption{LLM Inference Prompt Strategy Labels and Descriptions.}
    \vspace{.5em}
    \begin{small}
    \begin{tabular}{l@{\hskip .1in}l}
    \toprule
    \textbf{Label} & \textbf{Description}\\
    \midrule
    BFQH & Behavioral dynamics, free-text state description, reasoning questions, trajectory history\\  
    BFQ & Behavioral dynamics, free-text state description, reasoning questions\\ 
    BF  & Behavioral dynamics, free-text state description\\
    \bottomrule
    \end{tabular}
    \label{tab:LLM prompt ablation}
    \end{small}
    \end{center}
    \vspace{-2em}
\end{table}
}

\subsection{StepCountJITAI+LLM}
\label{sec: create StepCountJITAI for LLM}

We next turn to the design of a simulation environment to evaluate the proposed approach. We extend the base simulator introduced in \citet{karine2024}  and described in Section \ref{sec: background StepCountJITAI} to create an enhanced physical activity messaging-based intervention simulator that generates participant state descriptions using an additional LLM component. We describe each enhancement to the simulator below.

\vspace{.5em}

\label{Section: Creating  auxiliary variable W (cannot walk / can walk)}
\noindent\textbf{State Augmentation}. We introduce a new binary state variable $w_t\in\{0,1\}$ indicating whether the participant is able to engage in walking or not at time $t$. We use a Markov chain to simulate $w_t$. The Markov chain is illustrated  in Figure \ref{fig: Markov chain sketch} and Table \ref{tab: Transition Function}.  We parameterize the Markov chain  via $ p_{w_{00}} = P(w_{t+1}=0|w_t=0)$, the probability of staying in the ``can't walk" state, and $p_{w_{11}} = P(w_{t+1}=1|w_t=1)$, the probability of staying in the ``can walk" state. All simulations begin with participant in the ``can walk" state. The parameters of this Markov chain can then be set to simulate a participant that has a larger or smaller chance of becoming unable to walk, and how long the participant takes on average to recover after they become unable to walk. In our experiments, we consider several different scenarios based on different settings of these parameters.

\begin{minipage}{.9\textwidth}
    \begin{minipage}[b]{0.49\textwidth}
    \begin{figure}[H]
        \begin{center}            
            \begin {tikzpicture}[scale=0.55, ->, line width=0.5 pt, node distance=1.3cm]
                \node[circle,draw,thick,fill=gray,text=white](zero) {0};
                \node[circle,draw,thick,fill=gray,text=white](one) [right of=zero] {1};
                \path(zero) edge [loop left,red] node {$p_{w_{00}}$} (zero);
                \path(zero) edge [bend left,red] node[above] {$1-p_{w_{00}}$} (one);
                \path(one)  edge [bend left,blue] node[below] {$1-p_{w_{11}}$} (zero);
                \path(one)  edge [loop right,blue] node {$p_{w_{11}}$} (one);
            \end{tikzpicture}
        \end{center}
        \vspace{-1em}
        \caption{Markov chain sketch.}
        \label{fig: Markov chain sketch}
       \end{figure}    
    \end{minipage}
    \hfill
    \begin{minipage}[b]{0.49\textwidth}  
    \begin{table}[H]
    \begin{small}
    \caption{Transition Function.}
    \vspace{.5em}
    \label{tab: Transition Function}
    \centering
        \begin{tabular}{@{\hskip .2in}c@{\hskip .2in}|@{\hskip .2in}c@{\hskip .2in}||@{\hskip .2in}c@{\hskip .2in}}
            \toprule
            $w_t$ & $w_{t+1}$ & $P(w_{t+1}|w_t)$ \\
            \hline
            0 & 0 & $p_{w_{00}}$ \\
            0 & 1 & $1-p_{w_{00}}$ \\
            1 & 0 & $1-p_{w_{11}}$ \\
            1 & 1 & $p_{w_{11}}$ \\
            \bottomrule
        \end{tabular}
    \end{small}
    \end{table}
    \end{minipage}
\end{minipage}

\vspace{2em}


\begin{figure}[t!]
\noindent\fcolorbox{black}{black!2}{\begin{minipage}{.97\columnwidth}
\begin{footnotesize}
\noindent\textsf{\textbf{Example ``can walk" state descriptions:} \textcolor{blue}{I am feeling good,
I'm in a great mood,
I feel energized,
I'm feeling positive,
I'm doing well today,
I feel great,
I'm in high spirits,
I feel focused,
I'm feeling relaxed,
I feel motivated,
I'm doing fine,
I feel optimistic,
I'm feeling calm,
I feel balanced,
I'm feeling strong,
I feel productive,
I'm in a positive state of mind,
I feel healthy,
I feel confident,
I feel alert,
...
}}
\end{footnotesize}
\end{minipage}}

\vspace{0.5em}

\noindent\fcolorbox{black}{black!2}{\begin{minipage}{.97\columnwidth}
\begin{footnotesize}
\noindent \textsf{\textbf{Example ``can't walk" state descriptions:} \textcolor{blue}{I am tired,
I do not want to walk,
I got an injury,
I have a headache,
My legs are sore,
I twisted my ankle,
I’m feeling dizzy,
I’m out of breath,
I have a cold,
I’m feeling weak,
I pulled a muscle,
My knee hurts,
I have blisters,
I feel nauseous,
I have stomach cramps,
I can’t find my shoes,
I don’t have time,
I’m waiting for someone,
It’s too hot outside,
It’s too cold outside,
...
}}
\end{footnotesize}
\end{minipage}}
    \caption{Examples of generated participant supplied state descriptions.}
    \label{fig:state descriptions}
\end{figure}

\noindent\textbf{Participant Provided State Description Generation}. We generate participant provided state descriptions conditioned on the variable $w_t$ using two different LLM prompts. When transitioning from $w_{t}=1$ to $w_{t+1}=0$, we emit text produced by prompting the LLM to generate a short description of a reason why a person might not be able to walk. When staying in the ``can't walk" state, no additional state descriptions are generated and the most recent state description stays active. When transitioning from $w_{t}=0$ to $w_{t+1}=1$, we emit text produced by prompting the LLM to generate a message describing that the participant is ``feeling fine." When staying in the ``can walk" state, we decide whether or not to emit a new participant supplied state description independently at each time point with probability $0.3$. This simulates realistic response rates for such interactions. 

To enable reproducible experiments, we use an LLM to pre-generate lists of simulated participant state descriptions consistent with each state of $w_t$. Specifically, we generate 500 state descriptions of each type using ChatGPT \citep{achiam2023gpt}. Examples of state descriptions generated for each condition are shown in Figure \ref{fig:state descriptions}. 

\vspace{0.5em}
\noindent\textbf{Behavioral Dynamics}. In addition to dynamics for the $w_t$ variable, we need to consider how the ``can/can't walk" state should interact with the other state variables already present in the simulator. Below we give the full behavioral dynamics equations  for the StepCountJITAI+LLM simulator (new terms are shown in \textcolor{blue}{blue}) followed by a brief explanation. 

\label{Inserting new constraints to impact behavioral dynamics}
\begin{small}
\begin{align}
        c_{t+1} &\sim \mathit{Bernoulli}(0.5), \;\;\; x_{t+1} \sim \mathcal{N}(c_{t+1}, \sigma^2), \;\;\; 
        p_{t+1} = P(C=1|x_{t+1}), \;\;\; l_{t+1} = p_{t+1} >0.5\\[6pt]
        h_{t+1} &=   \begin{cases}
                    (1-\delta_h) \cdot  h_{t}             &\text{~~~~~~~~~~~~~~~~~~if~} a_{t} = 0\\
                    \text{min}(1, h_{t} + \epsilon_h)     & \text{~~~~~~~~~~~~~~~~~~otherwise}\\
                \end{cases}\\[6pt]
        d_{t+1} &=   \begin{cases}
                    d_{t}   &\text{~if~} a_{t} = 0  \textcolor{blue}{\text{~and~} w_t=0 \text{~or~} 1}  \\
                    %
                    %
                    (1-\delta_d) \cdot  d_{t}             &\text{~if~} a_{t} \in \{1,c_{t}+2\} \textcolor{blue}{\text{~and~}}  \textcolor{blue}{~w_t=1 \text{~~(can walk)~}}  \\
                    %
                    %
                     \textcolor{blue}{\text{min}(1, d_t + \eta_d)}   &\textcolor{blue}{\text{~if~} a_{t} \in \{1,c_{t}+2\} \text{~and~} } \textcolor{blue}{~w_t=0 \text{~~(can't walk)~}} \\
                    %
                    %
                    \text{min}(1, d_{t} + \epsilon_d \textcolor{blue}{+ (1-w_t)  ~ \eta_d})   &\text{~otherwise}
                \end{cases}\\[6pt]
    z_{t+1} &=  \begin{cases}
                    m_{s}    + (1-h_{t+1}) \cdot  \rho_1  &\text{~~~~~~~~~~~if~} a_{t} = 1  \textcolor{blue}{\text{~and~} w_t=1 \text{~~(can walk)~}} \\
                    m_{s}    + (1-h_{t+1}) \cdot  \rho_2  &\text{~~~~~~~~~~~if~} a_{t} = c_{t}+2  \textcolor{blue}{\text{~and~} w_t=1 \text{~~(can walk)~}} \\
                    m_{s} ~ \textcolor{blue}{w_t}   & \text{~~~~~~~~~~~otherwise}
                \end{cases}
\end{align}
\end{small}

As in the base simulator, we use $c_t$ to represent a binary context. Habituation increases by additive increments of $\epsilon_h$ up to a maximum value of $1$ whenever a message is sent and decays by a multiplicative factor of $(1-\delta_h)$ only when a message is not sent. There is no interaction between the ``can/can't walk" state and habituation level. When the participant can walk, the step count $z_t$ depends on the action $a_t$, the context $c_t$  and the habituation level $h_t$. When the participant cannot walk, the step count is set to $z_t=0$. 

In the base simulator, the dynamics of disengagement risk are primarily driven by the accuracy of message tailoring (e.g., whether the selected message type matches the current context $c_t$). The basic intuition is that incorrect tailoring causes the risk of disengagement to increase (for example, due to loss of trust in the intervention system). We similarly model the effect of sending messages when the participant is in the ``can't walk" state as a tailoring error that causes disengagement risk to increase. The updated dynamics for $d_t$ are a function of the  action selected $a_t$, the context $c_t$, and walking state $w_t$. 

Specifically, if no message is sent to the participant ($a_t=0$), the disengagement risk remains the same. If a correctly tailored message is sent and the participant can walk, the disengagement risk decreases multiplicatively by a factor of  $(1-\delta_d)$.  If a correctly tailored message is sent, but the participant cannot walk, the disengagement risk is incremented by a new parameter $\eta_d$ up to a maximum value of $1$. If an incorrectly tailored message is sent, the disengagement risk is incremented by an amount $\epsilon_d$ if the participant can walk, and by an additional amount $\eta_d$ if the participant cannot walk. 
As in the base simulator, if the disengagement risk reaches a value of $d_t=1$, we model the participant as dropping out of the study and the trial ends.

%% file: LLM_TS4_experiments_setup.tex
\section{Experiments}
\label{sec: LLM RL Experiments}

In this section, we describe experiments and results. We begin by presenting general experimental protocols. We then describe individual experiments and discuss results. 

\subsection{Experimental Protocols}
All experiments use the StepCountJITAI+LLM simulator. In all experiments, we set $\delta_h=0.1$, $\epsilon_h=0.05$, $\delta_d=0.1$, $\epsilon_d=0.05$, $\rho_1=50$, $\rho_2=200$. This yields a modest increase in steps for providing correctly tailored intervention content. We give results in terms of excess steps above the baseline step count $m_s$. We vary the $p_{w_{11}}$, $p_{w_{00}}$, and $\eta_d$ parameters to simulate participants with different characteristics. The full state in all experiments corresponds to $s_t=[c_t,h_t,d_t,w_t]$. The observed state accessible to the base RL agent in all experiments is $\tilde{s}_t=[c_t, h_t, d_t]$. Thus, $w_t$ is latent from the perspective of the base RL agent. We simulate a trial of up to $50$ days with one decision point per day. The reward is set to $0$ for any time steps following a disengagement event. For both LLM4TS and TS, all learning occurs within-trial for an individual participant. There is no sharing of data between participants. We set the TS prior parameters to $\mu_{0a}=0$ and $\Sigma_{0a}=100I$ for each action $a$ and the reward noise variance  $\sigma_{Ya}^2 = 25^2$ for each action $a$ (see Equations \ref{equations:TS posterior1} and \ref{equations:TS posterior2}).

The participant provided state descriptions $f_t$ are a function of $w_t$ only as described earlier. We pre-generate a total of $1000$ state descriptions using ChatGPT, $500$ for each walking ability state. We sample from this set when running the simulator. For experiments using LLM inference, we evaluate Llama 3 70B, Llama 3 8B and Gemma 2 9B \citep{Llama2024,gemma2024}. We use the prompt components shown in Figure \ref{fig:prompt1} for all experiments. All learning experiments are repeated five times and results are given in terms of median performance along with the 25th and 75th percentiles.

\input{LLM_TS4.1_validate_generation}

\input{LLM_TS4.2_validate_inference}
\input{LLM_TS4.3_main_experiment}

\input{LLM_TS4.4_llm_runtime_metrics}

\input{LLM_TS4.5_action_selections}

\input{LLM_TS4.6_prompt_structures}

%% file: LLM_TS4.1_validate_generation.tex
\subsection{Validating LLM Generation}
\label{Sec: Validating LLM Generation}

The first question we pose is, are the ChatGPT generated descriptions consistent with the intended state values? To answer this question, we extract a data set at random from the $1000$ state descriptions containing $100$ total state descriptions including $50$ generated using the ``cannot walk'' prompt and $50$ generated using the ``feeling fine'' prompt. We blinded the generating labels and manually classified each state description. This evaluation showed that 100\% of the generated state descriptions were consistent with the intended state. This verifies that ChatGPT can generate descriptions consistent with the desired states.

%% file: LLM_TS4.2_validate_inference.tex
\subsection{Validating LLM Inference}
\label{Sec: Validating LLM Inference}

\begin{figure}[t!]
\begin{center}
\includegraphics[width=.24\columnwidth]{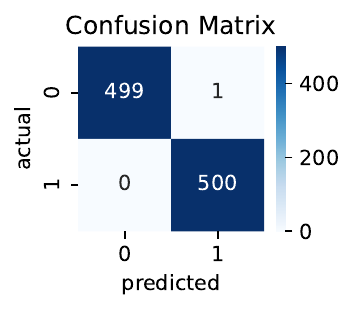}
\hfill
\includegraphics[width=.24\columnwidth]{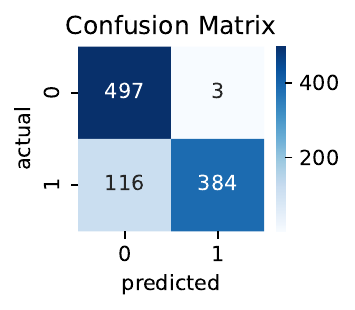}
\hfill
\includegraphics[width=.24\columnwidth]{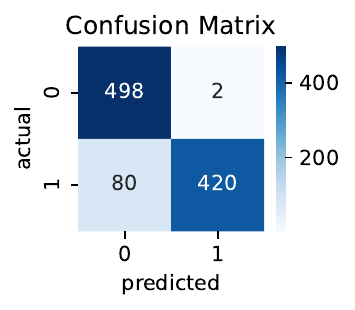}
\hfill
\vspace{-.5em}
\caption{Confusion matrices (left to right): Llama 3 70B, Llama 3 8B, Gemma 2 9B.
\vspace{-3em}}
\label{fig: Evaluating prompt strategy: BFQ 1000 data}
\end{center}
\end{figure}

The next question we ask is how well do different LLMs perform at the task of inferring whether messages should be sent based on simulated participant provided state descriptions using the developed prompt. We consider the true label to be ``send" when the walking state is ``can walk" and the true label to be ``don't send" when the walking state is ``can't walk." We evaluate Llama 3 70B, Llama 3 8B and Gemma 2 9B \citep{Llama2024,gemma2024}. We use an LLM temperature of $0.2$ (level of randomness in the LLM response) and the BFQ prompt strategy. We give confusion matrices for this inference problem in Figure \ref{fig: Evaluating prompt strategy: BFQ 1000 data}. We summarize accuracy, precision, recall and F1 metrics in Table \ref{tab: Validating LLM prompts}. We can see that while accuracy is positively correlated with model size among the three models tested, all three LLMs achieve accuracy above 85\%. For Llama 3 8B and Gemma 2 9B, we can see that the drop in accuracy is mostly accounted for by lower recall where these models incorrectly infer that messages should not be sent when sending messages is actually allowable.

\begin{table}[t!]
  \begin{center}
  \caption{Inference results using different LLMs.}
  \vspace{.5em}
  \label{tab: Validating LLM prompts}
  \begin{small}
  \begin{tabular}{c@{\hskip 0.3in}c@{\hskip 0.3in}c@{\hskip 0.3in}c@{\hskip 0.3in}c}
  \toprule
    \textbf{Model} & \textbf{Accuracy} & \textbf{Precision} & \textbf{Recall} & \textbf{F1}\\
    \midrule
    Llama 3 70B (BFQ) & 0.999 & 0.998 & 0.999 & 0.999 \\ 
    Llama 3 8B (BFQ)  & 0.881 & 0.992 & 0.881 & 0.866 \\ 
    Gemma 2 9B (BFQ)  & 0.918 & 0.995 & 0.918 & 0.911 \\ 
    \bottomrule
  \end{tabular}
  \end{small}
  \end{center}
\end{table}


%% file: LLM_TS4.3_main_experiment.tex
\subsection{Evaluating LLM4TS}
\label{sec: LLM4TS Experiments}

We conduct extensive experiments to compare LLM4TS to standard Thompson Sampling (TS). We explore four different scenarios for the dynamics of $w_t$. Scenario 1: $p_{w_{11}} = 0.7$, $p_{w_{00}}=0.5$. Scenario 2: $p_{w_{11}} = 0.7$, $p_{w_{00}}=0.1$, Scenario 3: $p_{w_{11}} = 0.95$, $p_{w_{00}}=0.5$. Scenario 4: $p_{w_{11}} = 0.95$, $p_{w_{00}}=0.1$. Among these scenarios, Scenario 1 has the highest per-time step chance of the participant becoming unable to walk, as well as the lowest chance of recovering. Scenario 4 has the lowest per-time step chance of the participant becoming unable to walk, as well as the highest chance of recovering. We also experiment with the value of $\eta_d$, the parameter controlling the disengagement risk increment when messages are sent in the ``can't walk" state. We consider the setting $\eta_d=0.05$, which corresponds to a participant whose disengagement risk increases very mildly in response to receiving messages in the ``can't walk" state, and $\eta_d=0.4$, which corresponds to a participant whose disengagement risk increases significantly.


\begin{figure}[t!]
\begin{center}
\includegraphics[height=0.165\textwidth]{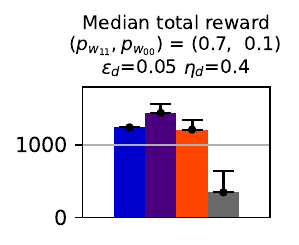}
\includegraphics[height=0.165\textwidth]{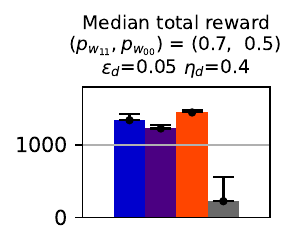}
\includegraphics[height=0.165\textwidth]{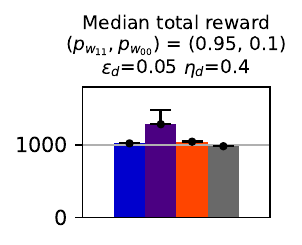}
\includegraphics[height=0.165\textwidth]{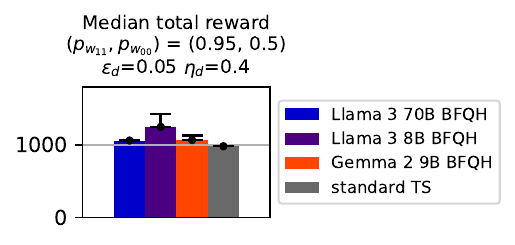}
%
\includegraphics[height=0.165\textwidth]{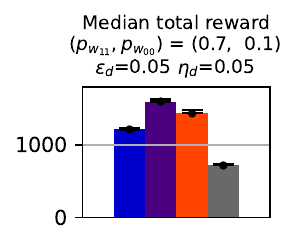}
\includegraphics[height=0.165\textwidth]{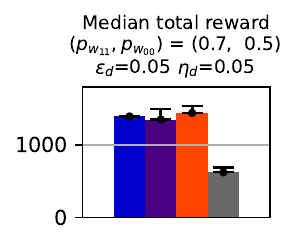}
\includegraphics[height=0.165\textwidth]{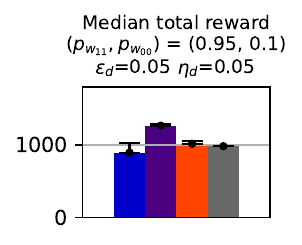}
\includegraphics[height=0.165\textwidth]{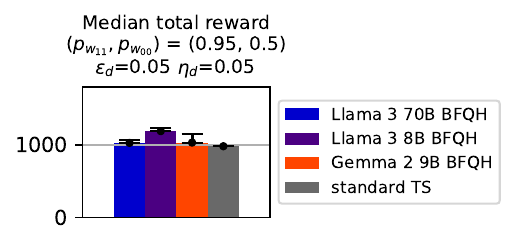}
\caption{Comparing LLMs. Columns correspond to different choices for $(p_{w_{11}}, p_{w_{00}})$. Rows correspond to different choices for  $\eta_d$. Bar colors indicate  different approaches.
\vspace{-3em}}
\label{fig: Comparing LLMS BFQH}
\end{center}
\end{figure}


\vspace{0.5em}
\noindent\textbf{Comparing LLM4TS and standard TS.} The question of interest in this experiment is under what circumstances can the  LLM4TS method improve on standard TS? We compare standard TS to LLM4TS using Llama3 70B, Llama3 8B and Gemma2 9B and the BFQH prompt. We show the results in Figure \ref{fig: Comparing LLMS BFQH}. Each row corresponds to a disengagement risk increment parameter value $\eta_d$. Each column corresponds to a different scenario for the $w_t$ dynamics. The bars correspond to different methods. The first trend we can see in these results is that when $p_{w_{11}}=0.7$, indicating a higher probability of transitioning to the can't walk state, LLM4TS exhibits significantly higher average reward than standard TS. This suggests that the LLM inference step and action filter are indeed contributing to improving adaptive intervention performance when participants are in the ``can't walk" state. 

Second, with $p_{w_{11}}=0.7$ we can observe a more subtle trend: the performance difference between LLM4TS and standard TS increases as the disengagement risk increment parameter $\eta_d$ is increased. This again makes sense as standard TS will trigger the disengagement risk threshold more often the higher the value of this parameter and therefore should obtain lower reward. The performance of LLM4TS does not vary significantly with $\eta_d$ since it is largely able to avoid sending messages in the ``can't walk" state.

Third, when $p_{w_{11}}=0.95$ indicating a low chance of transitioning to the ``can't walk" state, the performance of LLM4TS and standard TS are much closer regardless of the values of $\eta_d$ and $p_{w_{00}}$. This should be expected as LLM4TS has limited opportunities to improve performance when participants rarely enter the  ``can't walk" state. 

Lastly, we observe that when $p_{w_{11}}=0.7$, the total reward for LLM4TS is higher than when $p_{w_{11}}=0.95$. This is counter intuitive since the reward is zero when the participant can't walk. However, correctly inferring that a messages should not be sent when the participant can't walk results in the habituation level decreasing. Since the base TS agent selects messages to optimize expected immediate reward and the true expected immediate reward of sending a message is always higher than not sending a message, the base TS agent tends to send more messages than is optimal when taking into account the long term effect of actions. The LLM4TS action filter thus has a secondary positive effect due to decreasing habituation when standard TS is used as the base RL agent. Interestingly, this observation may also explain why Llama 3 70B has lower end-to-end performance despite having the best inference accuracy in the previous experiment.


%% file: LLM_TS4.4_llm_runtime_metrics.tex
\begin{figure}[t!]
\begin{center}
%
\includegraphics[height=0.165\textwidth]
{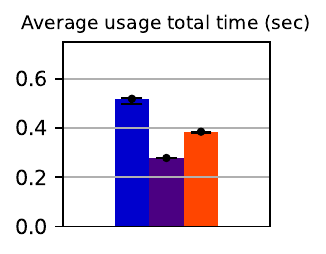}
\includegraphics[height=0.165\textwidth]
{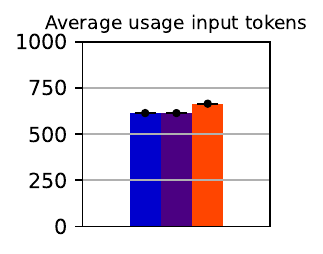}
\includegraphics[height=0.165\textwidth]
{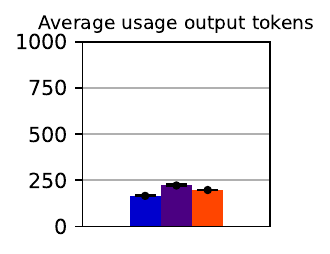}
\includegraphics[height=0.165\textwidth]
{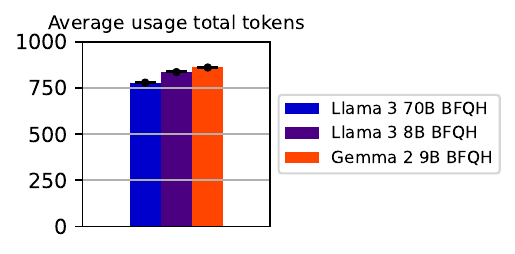}
\vspace{-1em}
\caption{Comparing LLM metrics for $\eta_d=0.4$ and $(p_{w_{11}}, p_{w_{00}})=(0.7,0.1).$
\vspace{-3em}}
\label{fig: Comparing LLMs metrics}
\end{center}
\end{figure}

\vspace{0.5em}
\noindent\textbf{LLM Runtime Metrics.} In this assessment, we examine the question of how runtime metrics differ for the LLMs used in the previous experiment. We examine  average per-call LLM inference time, average per-call input token count, and average per-call total (input+output) token count.  The results are shown in Figure \ref{fig: Comparing LLMs metrics}. We can see that the number of input tokens is similar for all LLMs as expected. However the average time per LLM inference call using LLama 3 8B is much lower than when using the other two LLMs while the average number of output tokens is only modestly higher than Llama 3 70B. However, the per-output token cost for LLama 3 8B is currently much lower than for LLama 3 70B on commercial LLM API providers, offsetting the higher output token count. These results combined with the results of the previous experiment indicate that LLama 3 8B offers strong performance in terms of average reward, inference time and cost. In the following experiments, we focus on Llama 3 8B.

%% file: LLM_TS4.5_action_selections.tex

\begin{figure}[t!]
\begin{center}
\includegraphics[width=.24\textwidth]{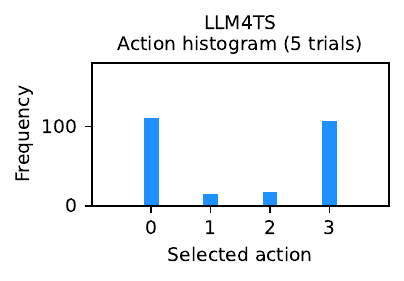}
\includegraphics[width=.24\textwidth]{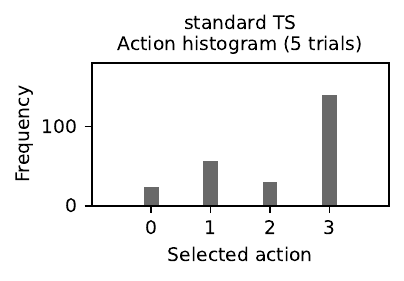}
\includegraphics[width=.24\textwidth]{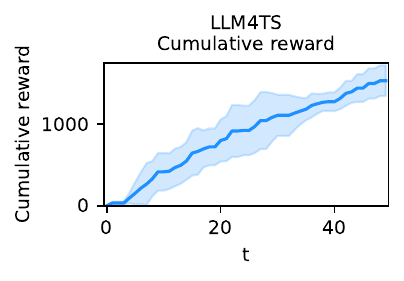}
\includegraphics[width=.24\textwidth]{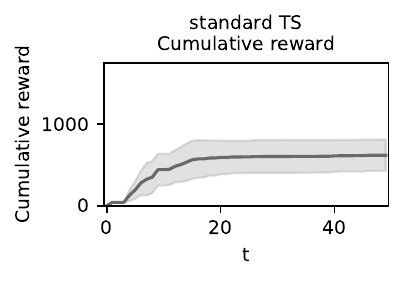}
\includegraphics[width=.24\textwidth]{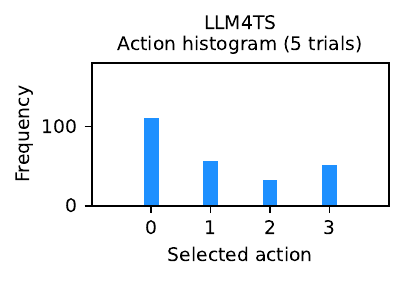}
\includegraphics[width=.24\textwidth]{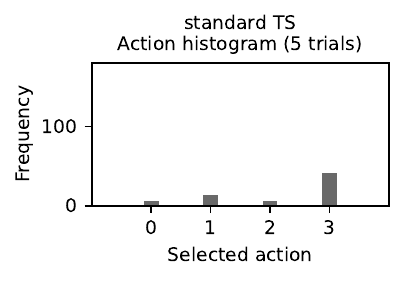}
\includegraphics[width=.24\textwidth]{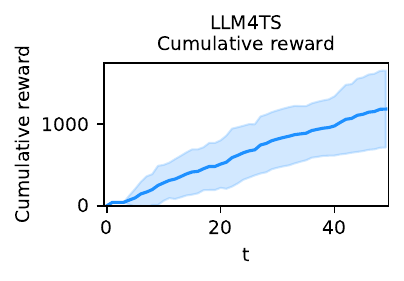}
\includegraphics[width=.24\textwidth]{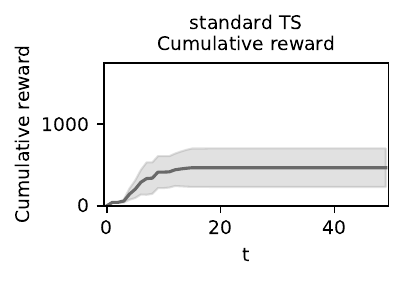}
\vspace{-.5em}
\caption{Histograms of action values and plots of average cumulative reward per episode for $(p_{w_{11}}, p_{w_{00}}) = (0.7, 0.1)$, with prompt strategy BFQH. (Top row) $\eta_d=0.05$. (Bottom row) $\eta_d=0.4$. (Blue) Llama 3 8B. (Gray) standard TS.\vspace{-2em}}
\label{fig: hist actions and returns medians pW11=0.7 Llama 3 8B BFQH}
\end{center}
\end{figure}

\vspace{0.5em}
\noindent\textbf{Analysis of Actions and States.} In this section, the primary question we address is the mechanism by which LLM4TS outperforms TS in different scenarios.
We first analyze the actions selected by LLM4TS compared to TS. We focus on $p_{w_{11}}=0.7$ and $p_{w_{00}}=0.1$ as a case where LLM4TS and TS differ in performance. We consider $\eta_d=0.05$ and $\eta_d=0.04$. We show results for Llama 3 8B in Figure \ref{fig: hist actions and returns medians pW11=0.7 Llama 3 8B BFQH}. The action selection histograms show that LLM4TS selects more $a_t=0$ actions, which indicates that LLM4TS has decided to not send a message more often than standard TS. We also compare the average cumulative reward per episode. The plateau seen in the TS cumulative reward plot suggests that TS is incurring disengagement events. 


\begin{figure}[t!]
\begin{center}
\includegraphics[width=0.49\textwidth]{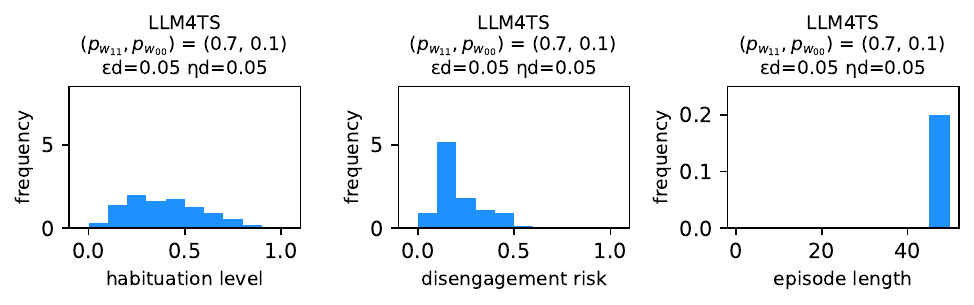}\hfill
\includegraphics[width=0.49\textwidth]{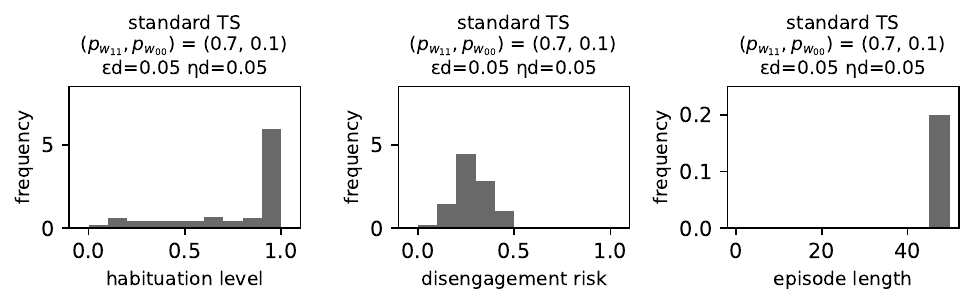}
\includegraphics[width=0.49\textwidth]{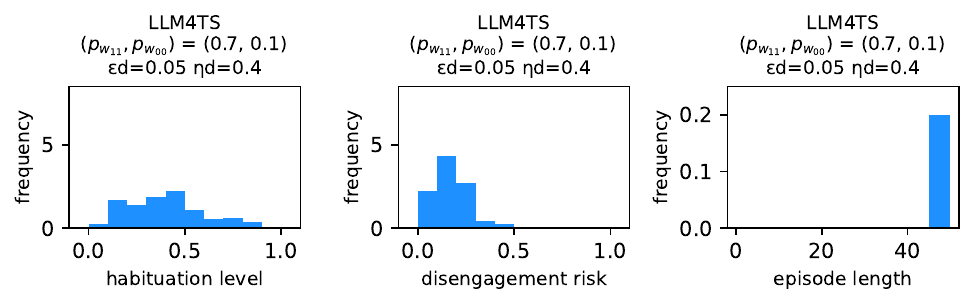}\hfill
\includegraphics[width=0.49\textwidth]{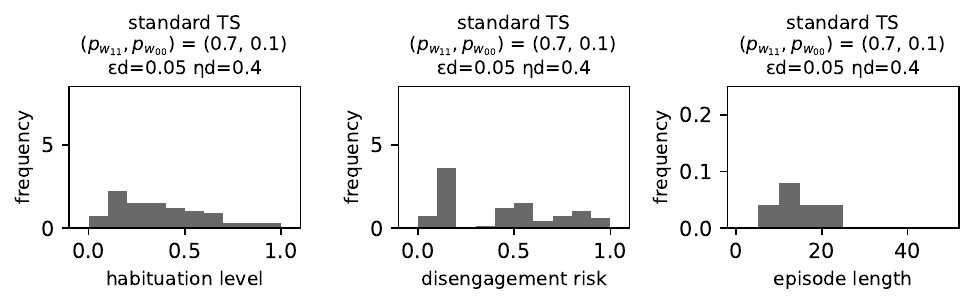}
\end{center}
\vspace{-1em}
\caption{Histograms of habituation level, disengagement risk, and episode length for $(p_{w_{11}}, p_{w_{00}}) = (0.7, 0.1)$, with prompt strategy BFQH. (Top row) $\eta_d=0.05$. (Bottom row) $\eta_d=0.4$. (Blue) Llama 3 8B. (Gray) standard TS.
\vspace{-1em}}
\label{fig: hist actions and returns scenarios Llama 3 8B BFH}
\end{figure}

Next, we analyze the distribution of habituation level, disengagement risk and episode length. We first note that for $\eta_d=0.05$, all trials for both methods complete all 50 time steps. This is possible for TS due to the low penalty on disengagement risk when sending messages while the participant cannot walk. However, we can see that the distribution of habituation values is concentrated on much higher values for TS than for LLM4TS, which results from sending messages during period where the participant cannot walk. This explains how LLM4TS outperforms TS in terms of total reward in this scenario. Next, when $\eta_d=0.4$, we can see that the distribution of habituation for plain TS is actually lower than for LLM4TS. However, the disengagement risk values are much higher and indeed we can see that none of the TS trials reach the full episode length indicating that all trials hit the disengagement risk threshold. Again, this can be attributed to TS sending messages when the participant cannot walk, while LLM4TS appropriately filters these actions.

%% file: LLM_TS4.6_prompt_structures.tex
\begin{figure}[t!]
\begin{center}
\includegraphics[width=0.21\textwidth]{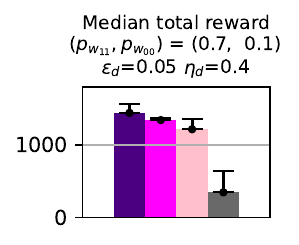}
\includegraphics[width=0.21\textwidth]{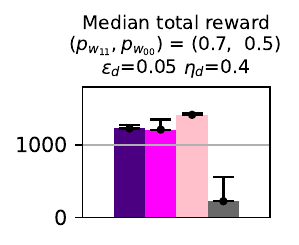}
\includegraphics[width=0.21\textwidth]{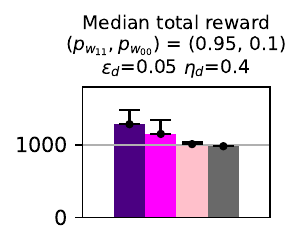}
\includegraphics[width=0.34\textwidth]{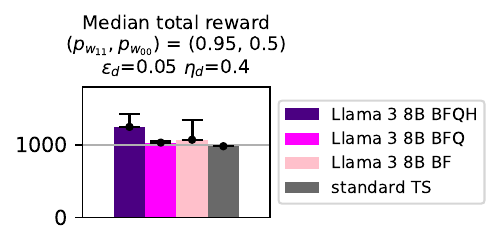}
\hfill
\includegraphics[width=0.21\textwidth]{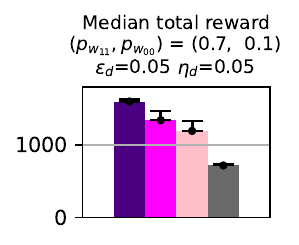}
\includegraphics[width=0.21\textwidth]{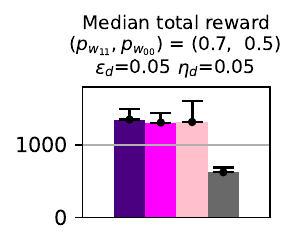}
\includegraphics[width=0.21\textwidth]{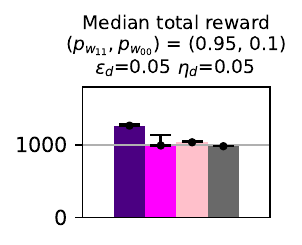}
\includegraphics[width=0.34\textwidth]{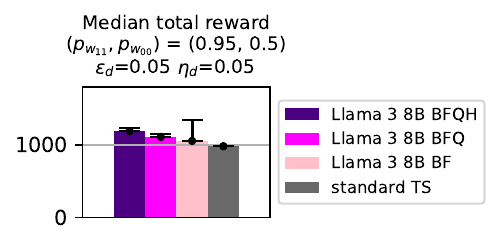}
\hfill
\end{center}
\vspace{-2em}
\caption{Comparing LLM prompt strategies, for various $(p_{w_{11}}, p_{w_{00}})$, using Llama 3 8B, with $\eta_d=0.4$ (top), $\eta_d=0.05$ (bottom). Standard TS is shown for comparison.}
\label{fig: Comparing LLM prompt strategies}
\end{figure}

\vspace{0.5em}
\noindent\textbf{Comparing Prompt Structures.} Lastly, we perform a comparison of different prompt structures. The primary question of interest is how does removing components from the BFQH prompt structure affect total reward? We consider the alternative prompt structures BFQ, and BF. The BFQH prompt structure corresponds to the example shown in Figure \ref{fig:prompt1}. The BFQ prompt structure removes the trajectory history. The BF prompt structure further removes the intermediate reasoning questions. We show the results for Llama 3 8B in Figure \ref{fig: Comparing LLM prompt strategies}. The columns again correspond to different choices for $p_{w_{11}}$, and $p_{w_{00}}$. We show results for $\eta_d=0.4$ and $\eta_d=0.05$. The results show that the BFQH prompt structure has median performance that is better than the next best prompt structure in six of the eight scenarios and is only outperformed in one scenario. These results suggest that performance degrades on average when removing components from the BFQH prompt structure.


%% file: LLM_TS5_conclusion.tex

\section{Discussion}
\label{sec: LLM RL Conclusion}

In this work, we have presented LLM4TS, an approach to augmenting a base RL method with LLM-based reasoning capabilities to help address the significant challenge of data scarcity that arises when applying RL methods to optimize adaptive intervention policies in the context of practical research study designs. Further, we have developed a physical activity JITAI simulation environment, StepCountJITAI+LLM, that models key behavioral dynamics as well as the generation of participant-provided state descriptions via an auxiliary LLM. We have presented experiments and results validating the generation of participant-provided state descriptions, the ability of LLMs to infer states from state descriptions, and the clear benefits of LLM4TS over standard Thompson Sampling in scenarios where the LLM reasoning component is exercised. These results support our claim that the LLM4TS approach is able to improve the limited state representation of a base Thompson Sampler while maintaining data efficiency.

We emphasize that the LLM4TS approach is a general and broadly applicable framework for enhancing the intelligence of adaptive interventions. The approach can be applied to different adaptive intervention domains by engineering appropriate LLM inference prompts. Our results suggest that supplying trajectory histories and intermediate reasoning questions along with participant provided state descriptions and hypotheses about behavioral dynamics contribute to improving performance. There is wide leeway for modifications such as supplying additional domain specific knowledge and investigating alternative intermediate reasoning questions. The approach can also be combined with any instruction tuned LLM and can thus benefit from future advances in LLM models. Similarly, while we have focused on applying this framework with a Thompson Sampler as the base RL method, it can be applied to any future advances in data efficient RL methods. 

\paragraph{Limitations.} While we believe our results show that the proposed approach holds significant promise for augmenting the intelligence of adaptive health interventions, this study has several limitations. First, our evaluation is limited to exploring the performance of the proposed approach in the context of a physical activity adaptive intervention simulation. While we expect LLMs to have sufficient world knowledge to provide appropriate reasoning in other behavioral intervention domains, this requires validation. For intervention domains where LLMs lack sufficient prior world knowledge, such knowledge could be provided as part of a reasoning prompt. Second, the utility of the approach hinges on the willingness of participants to provide state information to the intervention system. Our simulations do not assume that participants always provide responses when prompted, but real applications may need to contend with additional issues like informative missingness where participants are unlikely to respond when highly significant life events occur. Lastly, the magnitude of the performance improvements we observe in our experiments depends on many details of the simulation environment and we have highlighted multiple scenarios showing different levels of performance improvement. While validating this approach in simulation is an important first step, the next step for the approach requires evaluation in a human subjects study. The evidence presented in this paper will establish the foundation for conducting such a study. 

%% file: LLM_TS7_suppA.tex
\section{StepCountJITAI simulation environment}
\label{Appendix: StepCountJITAI}

The base simulator introduced in \citet{karine2023, karine2024} models the behavior of a participant in a messaging-based mobile health study. The intervention options (actions) are the messages sent to the participant, with the goal of increasing the participant walking step count (reward), given the participant's context (states). We summarize the base simulator specifications in Tables \ref{tab:env state} and \ref{tab:actions}, and provide details below.

\vspace{.5em}

\subsection{StepCountJITAI specifications}
\label{Appendix: StepCountJITAI simulation environment specifications}

For notation, we use an uppercase letter for the variable name, and a lowercase letter for the variable value. For example: the context variable $C$ has value $c_t =0$ at time $t$.
We describe some of the simulation environment variables and parameters that are used in the behavioral dynamics:
$c_t$ is the true context, $p_{t}$ is the probability of context $1$, $l_t$ is the inferred context, $h_t$ is the habituation level, $d_t$ is the disengagement risk, $z_t$ is the participant's walking step count, and $a_{t}$ is the action at time $t$. The base simulator also includes behavioral parameters: $\delta_d$ and $\epsilon_d$ are decay and increment parameters for the disengagement risk, and $\delta_h$ and $\epsilon_h$ are decay and increment parameters for the habituation level.
The \textbf{goal is to maximize the total walking step count}. Thus, the \textbf{walking step count is also the RL reward} ($r_t = z_{t}$). 

\begin{table}[ht]
    \begin{center}
    \small
    \caption{Environment state variables}
    \label{tab:env state}
    \vspace{.5em}
    \begin{tabular}{c@{\hskip 0.3in}l@{\hskip 
    0.3in}c}
    \toprule
        \bfseries Variable & \bfseries Description  & \bfseries Values\\
    \midrule
        $c_t$     & True context                  & $\{0,1\}$ \\
        $p_t$ & Probability of context $1$    &  $[0,1]$\\
        $l_t$     & Inferred context           & \{0,1\}\\
        $d_t$     & Disengagement risk level      & $[0,1]$\\
        $h_t$     & Habituation level             & $[0,1]$\\
        $z_{t}$     & Walking step count               & $\mathbb{N}$\\[1pt] 
    \bottomrule
    \end{tabular}
    \end{center}
    \vspace{-.5em}
\end{table}

\begin{table}[ht]
    \begin{center}
    \small
    \caption{Possible action values}
    \vspace{.5em}
    \label{tab:actions}  
    \begin{tabular}{c@{\hskip 0.3in}l}
        \toprule
            \textbf{Action} & \textbf{Description} \\ 
        \midrule
            $a_t=0$    &  No message is sent to the participant. \\[1pt]
            $a_t=1$    &  A non-contextualized message is sent. \\[1pt]
            $a_t=2$    &  A message customized to context $0$ is sent. \\[1pt]
            $a_t=3$    &  A message customized to context $1$ is sent. \\[1pt]
        \bottomrule
    \end{tabular}
    \end{center}
\end{table}

\subsection{StepCountJITAI behavioral dynamics}
\label{Section Background: behavioral dynamics}
The behavioral dynamics of the base simulator are as follow: Sending a message causes the habituation level to increase. Not sending a message causes the habituation level to decrease. An incorrectly tailored message causes the disengagement risk to increase. A correctly tailored message causes the disengagement risk to decrease. When the disengagement risk exceeds a given threshold, the behavioral study ends. The reward is the surplus walking step count, beyond a baseline count, attenuated by the habituation level. 
These behavioral dynamics equations for the base simulator are provided below. $\sigma$ is the context uncertainty, $x_t$ is a context feature. $\sigma, \rho_1, \rho_2, m_s$ are fixed parameters. 
\begin{small}
\begin{align}
        c_{t+1} \sim \mathit{Bernoulli}(0.5), \;\;\; &x_{t+1} \sim \mathcal{N}(c_{t+1}, \sigma^2), \;\;\; 
        p_{t+1} = P(C=1|x_{t+1}), \;\;\; l_{t+1} = p_{t+1} >0.5\\
        %
    %
        h_{t+1} &=   \begin{cases}
                    (1-\delta_h) \cdot  h_{t}             &\text{~~if~} a_{t} = 0\\
                    \text{min}(1, h_{t} + \epsilon_h)     & \text{~~otherwise}\\
                \end{cases}\\
        d_{t+1} &=   \begin{cases}
                    d_{t}                                 &\text{~~if~} a_{t} = 0\\
                    (1-\delta_d) \cdot  d_{t}             &\text{~~if~} a_{t} \in \{1,c_{t}+2\}\\
                    \text{min}(1, d_{t} + \epsilon_d)     &\text{~~otherwise}
                \end{cases}
\end{align}
\end{small}
\begin{small}
\begin{align}
    z_{t+1} &=   \begin{cases}
                m_{s}    + (1-h_{t+1}) \cdot  \rho_1  &\text{~~~~~~~~~~if~} a_{t} = 1\\
                m_{s}    + (1-h_{t+1}) \cdot  \rho_2  &\text{~~~~~~~~~~if~} a_{t} = c_{t}+2\\
                m_{s}    & \text{~~~~~~~~~~otherwise}
            \end{cases}
\end{align}
\end{small}